\documentclass{article}

\usepackage{PRIMEarxiv}

\usepackage[utf8]{inputenc} 
\usepackage[T1]{fontenc}    
\usepackage{hyperref}       
\usepackage{url}            
\usepackage{booktabs}       
\usepackage{amsfonts}       
\usepackage{nicefrac}       
\usepackage{microtype}      
\usepackage{lipsum}
\usepackage{fancyhdr}       
\usepackage{graphicx}       
\usepackage{multirow}
\usepackage{subcaption}
\usepackage{amsmath}
\graphicspath{{media/}}     

\title{Dual Co-Train: Cross-Dataset Ultrasound Tongue Segmentation Under Extreme Data Scarcity \thanks{This is a preprint}}

\author{
  Alisher Myrgyyassov\\
  Department of Biomedical Engineering\\
  Hong Kong Polytechnic University\\
  Hong Kong, China \\
  \texttt{24067038r@connect.polyu.hk} \\
   \And
  Zhen Song\\
  Department of Biomedical Engineering\\
  Hong Kong Polytechnic University\\
  Hong Kong, China \\
  \texttt{zhen0212.song@connect.polyu.hk} \\
  \And
  Bruce Xiao Wang\\
  Department of English and Communication\\
  Hong Kong Polytechnic University\\
  Hong Kong, China \\
  \texttt{brucex.wang@polyu.edu.hk} \\
  \And
  Yu Sun\\
  Department of Biomedical Engineering\\
  Hong Kong Polytechnic University\\
  Hong Kong, China \\
  \texttt{stefanie.sun@polyu.edu.hk} \\
  \And
  Min Ney Wong\\ 
  Department of Language Science and Technology \\
  Hong Kong Polytechnic University\\
  Hong Kong, China \\
  \texttt{min.wong@polyu.edu.hk} \\
  \And
  Yihao Zhou\\
  Department of Biomedical Engineering\\
  Hong Kong Polytechnic University\\
  Hong Kong, China \\
  \texttt{yihao.zhou@connect.polyu.hk} \\
  \And
  Yongping Zheng \\
  Department of Biomedical Engineering, Research Institute for Smart Ageing\\
  Hong Kong Polytechnic University\\
  Hong Kong, China \\
  \texttt{yongping.zheng@polyu.edu.hk} \\
}

\begin{document}
\maketitle

\begin{abstract}
Ultrasound tongue contour segmentation remains challenging under cross-dataset domain shift, where limited annotations, probe variability, and acquisition noise often degrade model generalization. We present a source-free domain adaptation framework for robust ultrasound tongue segmentation built on a lightweight UltraUNet backbone. Starting from a checkpoint pretrained on only five labeled source images, simulating an underfitted constrained source model, the proposed method adapts to a fully-unlabeled target domain by iteratively refining pseudo-labels, filtering unreliable masks with a contour-based quality-control module, and generating target-style synthetic image-mask pairs through a segmentation-guided conditional GAN. The student model is then trained on a mixture of clean pseudo-labeled target images, noisy pseudo-labels with consistency regularization, and synthetic samples, enabling closed-loop adaptation without access to source data. We evaluate the method on 12 source-target transfer pairs across eight ultrasound tongue imaging datasets, and conduct source-size scaling experiments and ablation studies. Across all comparisons, the proposed framework improves segmentation overlap and contour accuracy over the baselines, including supervised ones. These results suggest that task-specific pseudo-label refinement and synthetic target-style augmentation can substantially improve source-free adaptation for ultrasound tongue imaging.

\footnotetext{Code available at \url{https://github.com/AlisherMyrgyyassov/dual-co-train}}
\end{abstract}




\section{Introduction}

Ultrasound tongue imaging (UTI) provides a non-invasive and cost-effective means of visualizing tongue motion during speech, and it has become an important tool for articulatory phonetics, speech therapy, and the study of speech motor disorders \cite{utiReviewAlhammuri, zhuTongueSegmentation}. In practice, however, automated tongue contour segmentation remains challenging because UTI data are highly sensitive to probe placement, acquisition settings, operator skill, subject anatomy, and motion artifacts. These factors induce substantial domain shift across datasets, which limits the usefulness of models trained in one setting when deployed in another \cite{domainadaptationSurvey, ultraunet, medSegDAreview}.

Most of the proposed ultrasound tongue contour segmentation approaches were tested within single-dataset settings \cite{zhuTongueSegmentation, daftnet, feng2021tongueSegmentation, ultraunet, utiReviewAlhammuri} without true cross-domain evaluation on different datasets. Recent work on tongue contour segmentation, however, has demonstrated that deep learning can achieve strong cross-dataset performance, with UltraUNet \cite{ultraunet} emerging as a lightweight and efficient architecture for real-time UTI segmentation. The UltraUNet pipeline relies on ultrasound-specific augmentations, Squeeze-and-Excitation modules, and selective group normalization in deeper layers, as well as histogram matching for cross-dataset deployment. Although augmentation can improve robustness to some imaging variability, it typically does not directly address cross-dataset generalization, nor does it exploit the large amount of unlabeled ultrasound imaging data that is often available in practice. 

Utilizing these unlabeled datasets offers a highly promising alternative to mitigate domain shift, especially since expert annotation is labor-intensive and requires specialized domain knowledge, making large-scale labeling impractical \cite{domainadaptationSurvey, zhao2022unsupervisedDA, medSegDAreview}. To move beyond this regime, unsupervised domain adaptation (UDA) and, more recently, source-free domain adaptation (SFDA) have been investigated for medical image segmentation \cite{sffda_segmentation,sffda_medseg,medSegDAreview,sfdabaseline}. SFDA is particularly relevant in clinical and multi-site scenarios where only a trained source-domain model and unlabeled target images are available at adaptation time due to privacy and data-sharing constraints \cite{cityu-sfda}. In parallel, recent studies have shown that carefully designed few-shot fine-tuning of a source-pretrained model on a handful of labeled target images can be surprisingly competitive with SFDA in some settings \cite{fewshot_sfda}. However, both SFDA and few-shot adaptation have barely been explored for UTI, especially under realistic low-label conditions and severe equipment-induced domain shift.

In this paper, we address the problem of cross-dataset UTI segmentation under limited annotation from a source-free unsupervised domain adaptation perspective. Unlike traditional SFDA methods that often assume well-optimized source models, we intentionally initialize our framework with an UltraUNet trained on only five samples and then adapt this model to a new target dataset using only unlabeled target-data images. By doing so, we demonstrate that our framework, combining pseudo-labeling, segmentation-guided synthetic augmentation, and GAN–segmenter co-training, does not merely preserve the source knowledge but can also correct and generalize a highly constrained initial model under various domain shifts. This setting reflects realistic deployment scenarios in which adaptation to a new recording setup must be performed using only unlabeled target data, despite potentially large differences in transducers, acquisition protocols, and speaker populations.

To evaluate the method under realistic transfer conditions, we construct a benchmark covering eight datasets with substantial variability in acquisition conditions, including acoustic shadows, speckle noise, motion blur, and other artifacts. We consider multiple source–target transfers that span different speakers, recording setups, and ultrasound transducers, and we compare our approach against strong source-free adaptation baselines. We further perform ablation studies to quantify the contribution of each component of the proposed pipeline. Overall, our goal is to transform a within-dataset segmentation model into a reusable UTI segmentation framework that can be conveniently transferred from the low-label source dataset, reflecting the requirements of real-world deployment across sites and devices as well as ethical concerns.

The main contributions of this work are as follows:
\begin{itemize}
    \item We formulate ultrasound tongue contour segmentation as a low-label source-free cross-dataset adaptation problem and study transfer across eight heterogeneous UTI datasets under realistic imaging variability.
    \item We propose a source-free unsupervised adaptation framework that combines pseudo-labeling, contour-based quality control, conditional GAN-based synthetic augmentation, and GAN-segmenter co-training for efficient target-domain refinement.
    \item We demonstrate that a lightweight real-time segmentation model can be pretrained from only five labeled images in the base domain and subsequently adapted to a new target domain using unlabeled target frames without any additional annotations, making the framework practical for UTI research and speech applications.
    \item We validate the method through cross-dataset transfer experiments, comparisons with supervised and source-free baselines, and ablations that show the effect of each design choice on generalization performance.
\end{itemize}

\section{Related Works}
\subsection{Deep Learning for Ultrasound Tongue Imaging}
Early work on ultrasound tongue imaging (UTI) focused on contour extraction and tracking using classical image processing and probabilistic models, including active contours, active appearance models, and higher-order Markov random fields \cite{utiReviewXia2024, utiReviewAlhammuri}. More recently, deep learning has been widely adopted for UTI analysis, with convolutional neural networks (CNNs) and encoder-decoder architectures achieving strong performance for tongue contour tracking and segmentation \cite{utiDLreview,wunet,mozaffari2020bownet}. CNNs in particular demonstrate significant potential in UTI segmentation due to their superior pixel-level accuracy \cite{feng2021contourExtraction} and initialization-free nature compared to traditional methods \cite{mozaffari2020bownet}.

Despite these advances, existing methods face three fundamental limitations. First, most UTI segmentation studies are trained and evaluated within a single dataset or recording environment, often using relatively small subject pools and limited cross-dataset testing with true held-out data \cite{utiDLreview, ultraunet}. This narrow evaluation scope fails to assess generalization across the diverse imaging conditions, subjects, ultrasound transducer models, and language backgrounds encountered in realistic settings \cite{utiDLreview}. Second, many proposed methods require massive quantities of annotated frames for training (usually around 2000-4000 annotated images \cite{wunet, mozaffari2020bownet, ultraunet, daftnet}), creating a substantial annotation burden that limits practical deployment. Third, despite the abundance of unlabeled UTI data, these resources remain largely unexploited in existing segmentation pipelines.

Transfer learning has been explored as a potential strategy to address the annotation problem by reusing feature extractors across speakers or corpora \cite{mozaffari2019transfer}. However, published evaluations typically involve modest domain shifts (e.g., related corpora or same imaging device) and do not systematically address large cross-dataset differences in hardware, acquisition protocols, and population characteristics \cite{utiReviewXia2024,utiDLreview, ultraunet}. This leaves two critical gaps: models trained with substantial supervision still struggle to generalize across substantially different UTI datasets, and the potential of unlabeled data to improve both data efficiency and cross-domain robustness remains largely unexplored.

In terms of cross-dataset generalization assessment, UltraUNet represents a recent step toward real-time UTI segmentation, offering a lightweight architecture and a multi-dataset benchmark that spans several corpora and acquisition conditions \cite{ultraunet}. The architecture is essentially a U-Net \cite{unet} adaptation optimized for high-framerate inference, incorporating selective lightweight squeeze-and-excitation blocks in the deeper layers of the encoder for channel-wise feature recalibration, alongside group normalization to preserve training stability during small-batch operations. UltraUNet emphasizes cross-dataset evaluation and investigates augmentation strategies to emulate domain variability, but its training protocol remains fully supervised and does not explicitly leverage unlabeled data for domain adaptation. As a result, UltraUNet still inherits the broader limitations of supervised UTI segmentation: dependence on expert annotations and incomplete treatment of domain shift.

\subsection{Semi-Supervised and Source-Free Domain Adaptation in Ultrasound and Medical Segmentation}

Semi-supervised learning (SSL) and self-supervised learning have attracted substantial attention in medical image analysis as strategies to reduce annotation requirements by exploiting unlabeled images \cite{sslSegReview2023, ssl4mis}. For segmentation, popular approaches include consistency regularization, pseudo-labeling, cross-teaching, and perturbation-based teacher-student frameworks \cite{meanteacher, luo2022crossTeaching, sslSegReview2023}. These methods have been applied to MRI, CT, and ultrasound, and they have shown strong potential for improving label efficiency when expert annotation is expensive or scarce.

However, these SSL-based methods typically assume access to the same domain or a closely related training distribution, and they do not directly address the source-to-target shift that arises in cross-dataset deployment. In this respect, source-free domain adaptation (SFDA) is more relevant to our setting, because it assumes that only a trained source model and unlabeled target data are available at adaptation time. Recent SFDA work in medical image segmentation has explored prototype alignment, pseudo-label refinement, teacher-student consistency, and synthetic target adaptation \cite{medSegDAreview, sffda_segmentation, sffda_medseg}. Nevertheless, SFDA remains largely underexplored for ultrasound tongue contour segmentation, particularly in low-label settings where the source model is pretrained on only a few annotated images and the target domain differs substantially in acquisition conditions.

In UTI specifically, unlabeled data have mainly been explored in self-supervised or reconstruction-based representation learning \cite{utiReviewXia2024} or cross-modal prediction \cite{liu2021utiToLips, xiong2022utiToPhon}, rather than in source-free contour segmentation. These studies indicate that large unlabeled UTI corpora can be informative for downstream learning, but they do not directly solve source-free contour adaptation under cross-dataset domain shift. This gap motivates the dual co-training framework proposed in this work.

\begin{figure}
    \centering
    \includegraphics[width=0.5\linewidth]{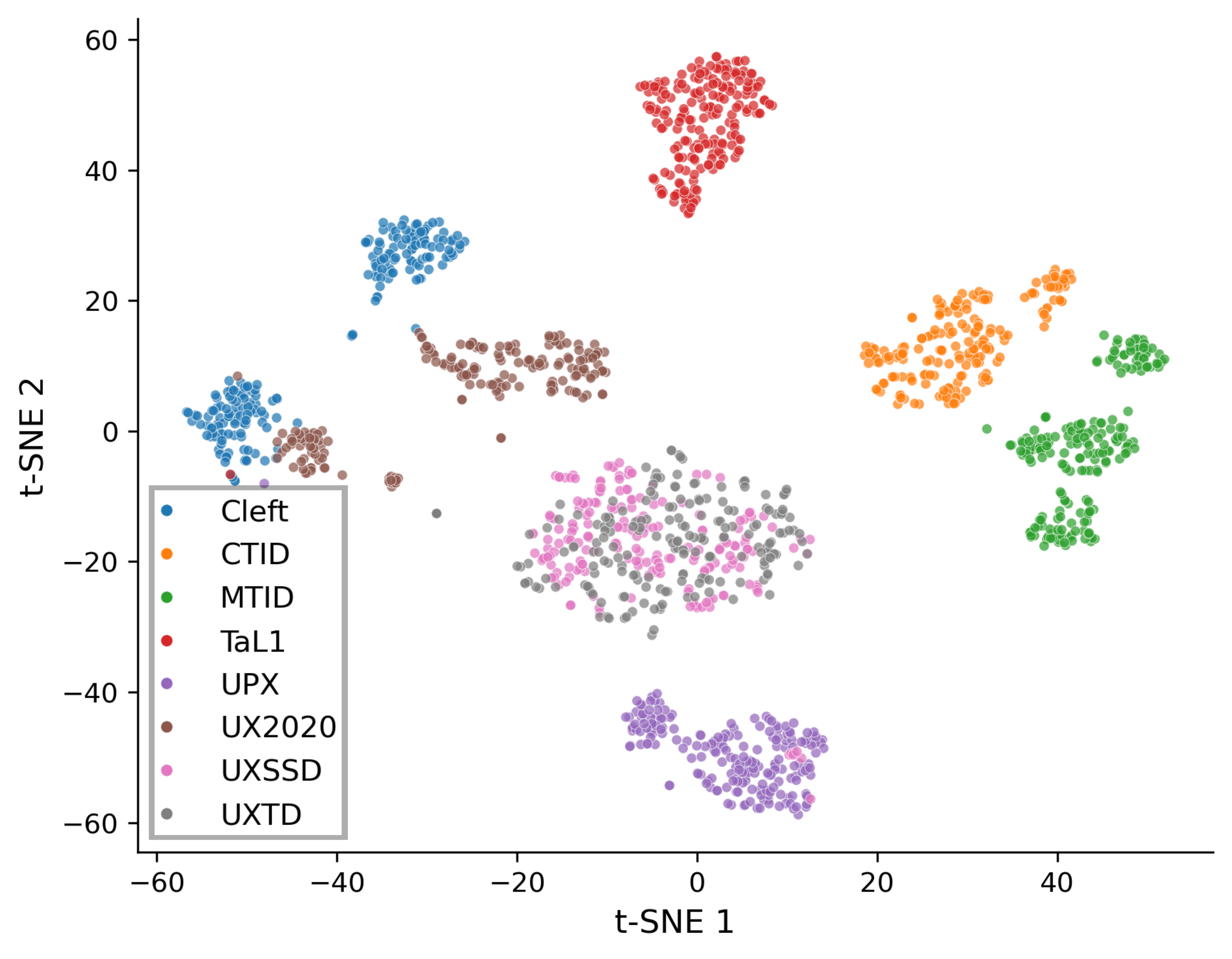}
    \caption{Two-dimensional embedding of image-level features from the eight ultrasound tongue imaging datasets extracted from the bottleneck of the pretrained UltraUNet, visualized using t-SNE. Each point corresponds to one 224$\times$224 image, and colors indicate dataset identity.}
    \label{fig:tsne}
\end{figure}

\section{Methods}
\subsection{Datasets}
We consider eight ultrasound tongue imaging test datasets, each comprising 200 annotated (using \cite{via}) frames sampled across all available speakers to cover inter-speaker anatomical variability as well as typical within-session variation in articulation. All images are cropped and resized to 224x224px. Datasets include newly annotated TaL1 \cite{talDataset}, UXTD, UXSSD, UPX, UX2020 \cite{eshky2018ultrasuite}, Cleft \cite{cleftDataset}, MTID, and CTID \cite{ultraunet}, following the held-out test setting, where the entire dataset remains completely unseen to the model. Frames are sampled from the ultrasound videos and reflect realistic clinical and research acquisition conditions, including speckle noise, acoustic shadowing, and motion blur. The resulting benchmark simulates a practical deployment scenario in which pretrained models are expected to generalize to new sites, scanners, and protocols. Fig. \ref{fig:tsne} shows the t-SNE embedding of bottleneck features extracted from a pretrained UltraUNet, with each point representing one 224$\times$224 image and colors indicating dataset identity to visualize the variability between and within datasets.

For each run within the experiment, one dataset is designated as the source domain with only 5 labeled images sampled from it for training. If a dataset is selected as the target dataset, its annotations will remain completely unseen to the model, and evaluation is performed on its full set of 200 images. For inter-rater analysis, 50 images per dataset are independently re-annotated by a second annotator, and 15 images per dataset are reannotated by a trained clinical sonographer. Inter-rater variability is then computed separately for each dataset. The sonographer trained both annotators before the annotation process.

To characterize image quality across the eight datasets, we computed a set of quantitative metrics (Table~\ref{tab:quality_tiers}). FG\% is the percentage of non-background pixels per frame, defined as pixels with non-zero intensity relative to all zero-intensity (background) pixels in the $224\times224$ image. Speckle SNR is $\mu_{tissue}/\sigma_{tissue}$ within the foreground. Norm. Gradient and Norm. Laplacian are the mean Sobel gradient and Laplacian variance within tissue, normalized by foreground mean intensity and its square, respectively.

\begin{table}[htbp]
\centering
\caption{Quantitative image-quality metrics for each dataset.}
\label{tab:quality_tiers}
\begin{tabular}{lcccc}
\toprule
Dataset & FG\% & Speckle SNR & Norm. Grad. & Norm. Lap. \\
\midrule
MTID  & 58.2 & 1.534 & 1.040 & 0.499 \\
CTID  & 60.1 & 1.531 & 1.023 & 0.279 \\
UXTD   & 52.9 & 1.010 & 2.069 & 1.144 \\
UXSSD  & 52.1 & 1.007 & 2.063 & 1.114 \\
TaL1   & 21.1 & 1.015 & 2.934 & 2.048 \\
UPX    & 54.0 & 0.902 & 1.611 & 0.696 \\
Cleft  & 22.7 & 0.933 & 2.828 & 2.512 \\
UX2020 & 39.7 & 0.748 & 2.898 & 3.669 \\
\bottomrule
\end{tabular}
\end{table}

\begin{table}[t]
    \centering
    \caption{Inter-rater MSD (in px) across the eight ultrasound tongue imaging datasets. A1 denotes the primary annotator, A2 denotes the secondary annotator, and S denotes the clinical sonographer. Values are reported as mean $\pm$ standard deviation.}
    \label{tab:msd_inter_rater}
    \begin{tabular}{lccc}
        \toprule
        Dataset & A1 vs A2 & A1 vs S & A2 vs S \\
        \midrule
        Cleft  & 2.583 $\pm$ 3.517 & 3.151 $\pm$ 2.334 & 3.112 $\pm$ 3.660 \\
        CTID   & 1.969 $\pm$ 1.456 & 1.518 $\pm$ 0.676 & 1.541 $\pm$ 0.272 \\
        MTID   & 2.225 $\pm$ 2.347 & 2.160 $\pm$ 2.989 & 1.802 $\pm$ 1.084 \\
        TaL1   & 3.861 $\pm$ 2.684 & 3.571 $\pm$ 2.607 & 3.682 $\pm$ 2.674 \\
        UPX    & 1.526 $\pm$ 0.828 & 1.182 $\pm$ 0.286 & 1.529 $\pm$ 0.120 \\
        UX2020 & 2.686 $\pm$ 1.858 & 2.418 $\pm$ 1.325 & 2.310 $\pm$ 0.852 \\
        UXSSD  & 1.370 $\pm$ 1.020 & 1.387 $\pm$ 0.909 & 1.033 $\pm$ 0.148 \\
        UXTD   & 2.080 $\pm$ 3.631 & 2.149 $\pm$ 1.792 & 1.966 $\pm$ 1.103 \\
        \bottomrule
    \end{tabular}
\end{table}

Table \ref{tab:msd_inter_rater} summarizes the inter-rater MSD across the eight datasets, revealing substantial variability in human contour agreement. The most challenging corpus is TaL1, where MSD between annotators ranges from 3.57 to 3.86px, indicating larger disagreement on tongue boundary placement. In contrast, UXSSD and UPX exhibit the lowest inter-rater MSD (around 1.0–1.5px), suggesting that tongue contours in these datasets are more consistently delineated across raters. These human variability ranges provide a practical reference band for interpreting model MSD on each target domain.

\begin{figure}
    \centering
    \includegraphics[width=0.95\linewidth]{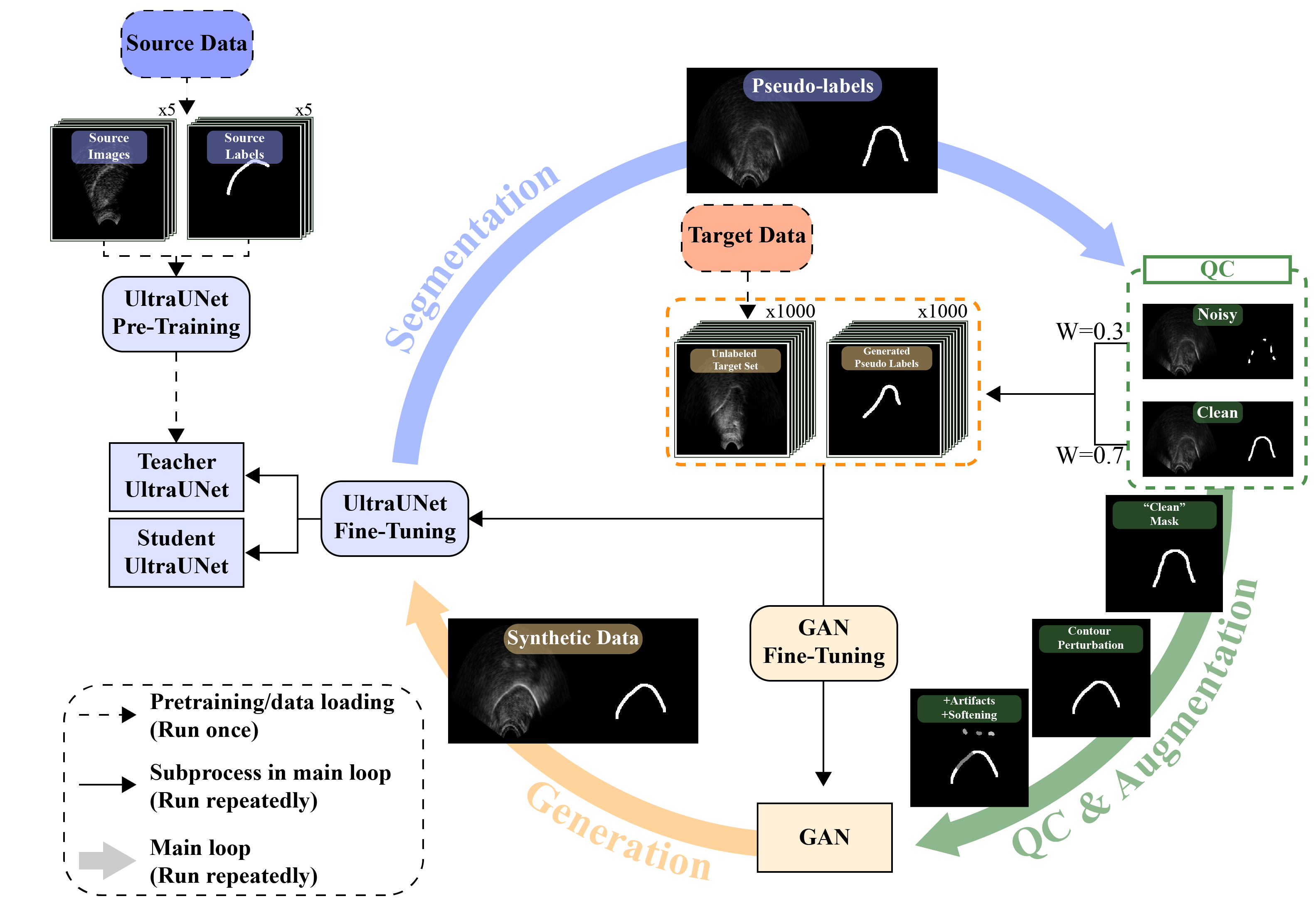}
    \caption{Detailed fine-tuning procedure. Teacher-generated pseudo-masks on unlabeled target images are separated into clean and noisy subsets by a rule-based quality-control module. Clean pseudo-labels and GAN-synthesized image-mask pairs are used for supervised segmentation training, while noisy pseudo-labels are handled through Mean Teacher consistency regularization. The GAN is periodically fine-tuned using updated pseudo-masks to remain aligned with the evolving target-domain mask distribution.}
    \label{fig:fine-tuning}
\end{figure}

\begin{figure}
    \centering
    \includegraphics[width=1\linewidth]{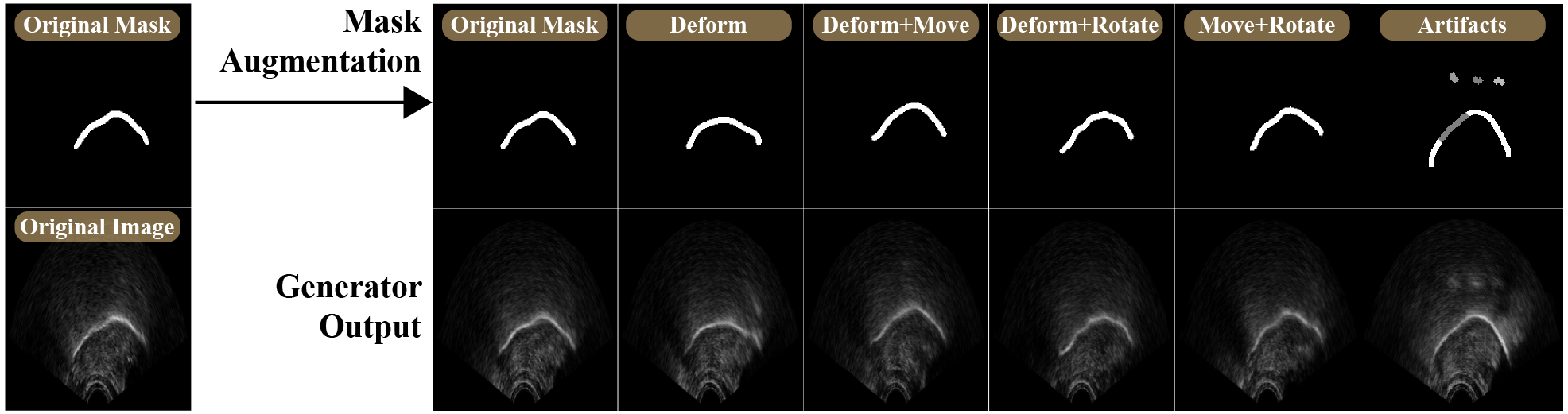}
    \caption{Mask augmentation pipeline. The original image is segmented to extract the input mask, which is then passed through the augmentation pipeline to generate new synthetic samples through deformation, moving, and rotation.}
    \label{fig:gans-augmentation}
\end{figure}

\begin{figure}
    \centering
    \includegraphics[width=1\linewidth]{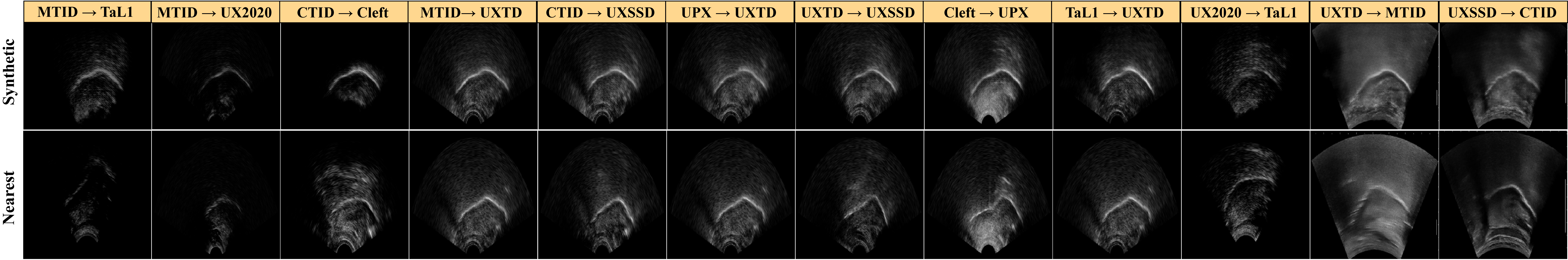}
    \caption{Comparison of synthetic and nearest real images generated using pretrained GANs from 12 transfer scenarios. (Top row) Aligned synthetic images generated from identical input masks. (Bottom row) The nearest corresponding real images, retrieved based on pixel value similarity (L2 distance).}
    \label{fig:gans-different}
\end{figure}

\subsection{Dual Co-Train Framework Overview}
The proposed method is a dual co-training framework that couples a segmentation network with a segmentation-guided conditional generative adversarial network (GAN) \cite{pix2pix}. The core idea is to adapt a source-trained tongue contour segmenter to an unlabeled target domain by iteratively refining pseudo-labels on real target images while simultaneously synthesizing target-style image--mask pairs that improve segmentation robustness.

The framework consists of three stages: 1) supervised source pretraining of the segmenter, 2) GAN pretraining on the pseudo masks generated from the pretrained segmenter, 3) target-domain fine-tuning through pseudo-label-driven dual co-training. The framework is evaluated using the held-out target testing without seeing any labels for the target domain. During fine-tuning, the segmenter produces pseudo-masks for unlabeled target images, a Quality-Control (QC) module separates reliable and unreliable pseudo-labels, and a conditional GAN is fine-tuned to synthesize target-style ultrasound images conditioned on pseudo-mask geometry. The segmentation model is then updated using a mixture of clean pseudo-labeled real images, noisy pseudo-labeled real images under consistency regularization, and synthetic image-mask pairs generated by the GAN.

To better illustrate the rationale behind the proposed Dual Co-Train framework, let $S_\phi$ denote the segmentation network with parameters $\phi$, and let $G_\theta$ denote the conditional GAN with parameters $\theta = (\theta_G, \theta_D)$ where $\theta_G$ and $\theta_D$ are the generator and discriminator parameters, respectively. We write $\mathcal{X}_t$ for the target image domain, $x_t \in \mathcal{X}_t$ for a target image sample, and $\hat{Y}_t^{(k)}$ for the pseudo-mask distribution estimated from the target domain at adaptation step $k$. The GAN learns a conditional mapping $G_\theta : (\hat{Y}_t, z) \rightarrow \tilde{X}_t$, where $z$ is a noise variable and $\tilde{X}_t$ denotes synthetic target-style images.

Because the teacher segmenter is updated during training, the induced pseudo-mask distribution $\hat{Y}_t^{(k)}$ is non-stationary. As the segmenter improves, the pseudo-mask geometry changes, and the GAN must be periodically fine-tuned so that its conditional image distribution remains aligned with the evolving mask distribution. In other words, the synthetic image generator must adapt whenever the pseudo-mask distribution shifts.

The closed-loop framework can be formulated explicitly as:
\[\hat{y}_t^{(k)} = S_{\phi_k}(x_t), \quad x_t \in \mathcal{X}_t\]
\[\tilde{x}_t = G_{\theta_k}(\hat{y}_t^{(k)}, z)\]

Then, after the segmenter parameters are updated to $\phi_{k+1}$, the GAN is refreshed using the newly induced pseudo-mask distribution.

\subsection{Source Segmentation Pretraining}
The segmentation backbone is UltraUNet \cite{ultraunet}, pretrained on a small set of five labeled source-domain images using strong paired augmentations applied to both images and contour-derived heatmaps. Training uses fixed-length epochs of 80 steps with resampling over the small labeled pool, for 20 epochs with batch size 2. The loss function is a weighted combination of focal loss and Dice loss, defined as:
\[
\mathcal{L}_{\mathrm{sup}} = 0.8\,\mathcal{L}_{\mathrm{focal}} + 0.2\,\mathcal{L}_{\mathrm{dice}}.
\]
The focal loss uses \(\alpha=0.25\) and \(\gamma=2.0\), while the Dice loss uses a smoothing constant of 1.0. Optimization is performed using AdamW \cite{adamw} (\(\mathrm{lr}=10^{-4}\), weight decay \(2\times10^{-4}\)) with gradient clipping at 1.5, and training is stopped early based on an EMA-smoothed training loss (patience 60). The resulting pretrained checkpoint is used to initialize the
subsequent source-free adaptation stage.

\subsection{Target-Domain Segmentation Fine-Tuning}
After source pretraining, the framework adapts the segmenter to the target domain using unlabeled target images. A student-teacher setup is employed, where the teacher network is maintained as an exponential moving average (EMA) of the student parameters:
\[
\theta_{\tau}^{(\mathrm{teacher})}
=
\lambda \theta_{\tau-1}^{(\mathrm{teacher})}
+
(1-\lambda)\theta_{\tau}^{(\mathrm{student})},
\]
where \(\lambda=0.99\) in all experiments.

For each segmentation epoch, the teacher first inferred pseudo-masks for the target unlabeled set using a probability threshold of 0.5. These pseudo-masks are then passed to a contour-based QC module that split them into clean and noisy subsets. The QC stage is rule-based and rejects masks exhibiting anatomically implausible or structurally unstable shapes, including tiny components, fragmented contours, border-touching masks, excessive holes, and broken skeletons, as summarized in Fig. \ref{fig:fine-tuning}.

The student is then optimized on a mixed training set composed of three branches:

\begin{enumerate}
    \item \emph{Clean} pseudo-labeled target images, trained with supervised segmentation loss;
    \item \emph{Noisy} pseudo-labeled target images, trained with a Mean Teacher consistency objective;
    \item \emph{Synthetic} target-style image--mask pairs generated by the GAN, trained with supervised segmentation loss.
\end{enumerate}

Accordingly, the total segmentation objective is defined as:
\[
\mathcal{L}_{\mathrm{seg}}
=
w_{\mathrm{clean}}\mathcal{L}_{\mathrm{sup}}^{\mathrm{clean}}
+
w_{\mathrm{synth}}\mathcal{L}_{\mathrm{sup}}^{\mathrm{synth}}
+
w_{\mathrm{noisy}}\mathcal{L}_{\mathrm{cons}},
\]
where \(w_{\mathrm{clean}}=0.7\), \(w_{\mathrm{synth}}=1.0\), and \(w_{\mathrm{noisy}}=0.3\). For noisy samples, \(\mathcal{L}_{\mathrm{cons}}\) is defined as the mean squared error between student and teacher sigmoid outputs under two independently augmented views of the same input. Target-domain segmentation fine-tuning is performed for 20 epochs.

\subsection{Segmentation-Guided GAN}
A Pix2Pix-style conditional GAN \cite{pix2pix} is used to synthesize target-style ultrasound images from mask-based conditioning. The generator receives a two-channel input formed by concatenating a contour mask and a Gaussian noise map, and produces a single-channel ultrasound image. The discriminator is a PatchGAN \cite{patchgan} operating on the concatenation of mask, image, and noise channels.

The GAN is first pretrained on the target small set for 30 epochs with a batch size of 8. Both the generator and discriminator used a learning rate of \(10^{-4}\). The generator objective combined adversarial,
\(L_1\), and perceptual losses: 
\[
\mathcal{L}_{G}
=
\mathcal{L}_{\mathrm{adv}}
+
\lambda_{1}\mathcal{L}_{1}
+
\lambda_{p}\mathcal{L}_{\mathrm{perc}},
\]
where \(\lambda_{1}=10\) and \(\lambda_{p}=8\) during GAN pretraining, selected via preliminary experiments. The perceptual term is computed using VGG19 \cite{vgg} features after replicating grayscale images to three channels \cite{perceptualLoss}. One-sided label smoothing is used in the discriminator, with real labels set to 0.9 and fake labels to 0.0, and Gaussian corruption with standard deviation 0.02
is applied to discriminator inputs.

During segmentation fine-tuning, the GAN is periodically updated to track the evolving target pseudo-mask distribution. Specifically, every four segmentation epochs, the teacher re-inferred pseudo-masks on the target adaptation set, and the GAN is fine-tuned for two epochs starting from the current best checkpoint. During this stage, the generator and discriminator learning rates are \(10^{-5}\) and \(10^{-4}\), respectively, the perceptual loss weight is increased to 20, and discriminator input noise is set to 0.05, all determined empirically. This closed-loop update is the key mechanism that keeps synthetic data aligned with the current target-domain contour distribution.

\subsection{Synthetic Sample Construction}
At each segmentation epoch, a synthetic pool is constructed from target pseudo-masks. To favor anatomically meaningful shapes, only the top 50\% of pseudo-masks ranked by foreground area are retained for synthetic generation. From this pool, 1000 synthetic samples are generated per epoch with replacement.

Utilizing GAN in the framework provides a variety of new augmentation techniques. Before GAN synthesis, each selected contour undergoes stochastic shape augmentation to increase geometric diversity. The augmentation operations include contour deformation, translation, and rotation, applied with probabilities of 0.6, 0.3, and 0.4, respectively (Figure \ref{fig:gans-augmentation}). Additional small artifacts are either retained from the original pseudo-mask or injected synthetically, each with a probability of 0.10. The augmented contour is then combined with Gaussian conditioning noise of amplitude 0.1 and passed through the GAN to generate a target-style ultrasound image. Synthetic images and the corresponding nearest real images from the pretrained GANs are shown in Figure \ref{fig:gans-different}.

After image synthesis, image-only augmentation is applied to better mimic target acquisition variability. This includes speckle corruption with a probability of 0.3 and noise standard deviation of 0.05, as well as point-spread-function blur with a probability of 0.3 and axial/lateral standard deviations sampled between 0.1 and 0.7. Global gain/bias augmentation and synthetic mask dilation are disabled in the final configuration.

\section{Experiments and Results}
\subsection{Cross-Dataset Adaptation Evaluation}

All models are evaluated on the held-out 200-image annotated test set of the target dataset. Predictions are thresholded at 0.5, and the largest connected component is retained before surface-based evaluation. Performance is quantified using mean sum distance (MSD) of the skeletonized mask as the primary tongue contour dissimilarity metric \cite{li2005automaticContourTracking, zhuTongueSegmentation} and Dice score to assess overlap between predicted and reference masks. Transfer pairs were selected using the quantitative image-quality metrics in Table \ref{tab:quality_tiers} to span the observed range of source-target domain-shift magnitude, from low-shift pairs between similarly clean datasets (e.g., MTID→CTID) to high-shift pairs spanning the largest quality gaps (e.g., CTID$\rightarrow$Cleft, MTID$\rightarrow$TaL1). 
We prioritized transfers originating from the two cleanest datasets (MTID, CTID) as sources, since prior work has shown this direction exposes generalization failure more severely than the reverse \cite{ultraunet}. Subject to this emphasis, pairs were chosen so that every dataset appears as a source at least once and as a target at least once, yielding 12 of the 56 possible ordered pairs, a subset chosen to keep evaluation computationally tractable across six methods and three random seeds while preserving representative domain-shift coverage. Each transfer experiment is repeated three times with different random seeds, following the outer-loop setup in the training pipeline, and all methods are compared under the same source-to-target transfer protocol.

\begin{figure}
    \centering
    \includegraphics[width=0.95\linewidth]{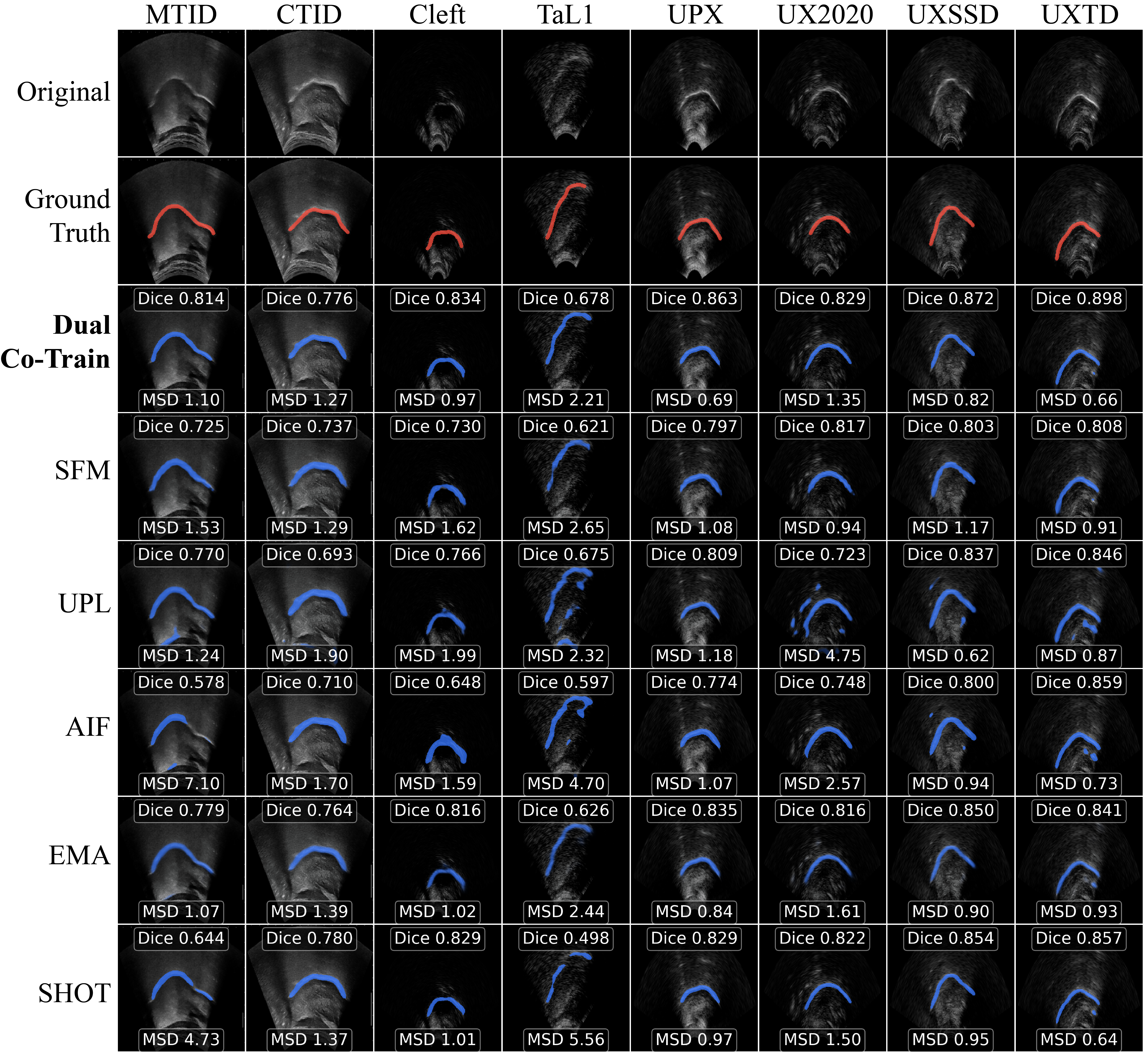}
    \caption{Representative qualitative comparison between 5 methodologies. Each column represents a dataset, and the rows are as follows: (1) original image, (2) human annotations overlay, (3) the proposed Dual Co-Train framework, (4) SFM baseline, (5) EMA teacher baseline, (6) SHOT baseline, (7) checkpoint trained on the target dataset}
    \label{fig:qualitative}
\end{figure}

We compare against five source-free baselines: a standard EMA teacher~\cite{meanteacher}, a Fourier Style Mining (FSM) framework~\cite{cityu-sfda}, a SHOT-style pseudo-label baseline~\cite{sfda-baseline-pseudo}, an Uncertainty-aware Pseudo Label (UPL) baseline~\cite{upl}, and an Autonomous Information Filter (AIF) baseline~\cite{aif}. The EMA teacher adapts UltraUNet directly via exponential moving averaging of the student parameters, without pseudo-label filtering or synthetic augmentation. FSM inverts target images into source-like images using BatchNorm statistics and Fourier domain adaptation, then adapts the model with compactness-weighted pseudo-label and contrastive distillation losses; following the original settings, our reimplementation uses a DeepLab backbone rather than UltraUNet, since the original version of UltraUNet uses selective GroupNorm instead of the required BatchNorm. The SHOT-style baseline retains only high-confidence pseudo-labeled pixels under an entropy-minimization regularizer and GroupNorm. UPL duplicates the decoder into four heads and builds confidence-thresholded ensemble pseudo-labels, supervised with reliability-masked Dice loss. AIF learns a DCT-based frequency filter and adapts via teacher-student EMA with CLUB mutual-information and cosine consistency on bottleneck embeddings. Qualitative results are shown in Fig.~\ref{fig:qualitative}.

Tables \ref{tab:dice_results} and \ref{tab:msd_results} report Dice and MSD for the 12 source-to-target transfer pairs, respectively. On average, the proposed dual co-training framework achieves the highest Dice (0.760) and lowest MSD (2.412\,px) among all six methods, outperforming EMA (0.723 Dice, 3.015\,px), SHOT (0.702, 3.537), UPL (0.693, 3.978), AIF (0.668, 4.147), and FSM (0.653, 4.345), which consistently underperforms the other baselines. Our method attains the best Dice in 10 of 12 pairs; the only exceptions are MTID$\rightarrow$TaL1, where EMA marginally leads (0.793 vs.\ 0.790), and TaL1$\rightarrow$UXTD, where UPL marginally leads (0.671 vs.\ 0.653), both involving comparatively noisy source or target datasets.

To assess whether these differences are statistically robust, we conducted a Friedman test across the six source-free methods (Ours, EMA, SHOT, UPL, AIF, FSM) over the 12 transfer pairs. The test revealed significant overall differences for both Dice ($\chi^2 = 38.81$, $df = 5$, $p < 0.001$) and MSD ($\chi^2 = 32.10$, $df = 5$, $p < 0.001$). Post-hoc pairwise Wilcoxon signed-rank tests with Bonferroni correction ($\alpha = 0.01$) confirmed that our method significantly outperformed all five baselines on both Dice and MSD ($p_{bonf} \leq 0.005$ in every comparison).

\begin{table*}[t]
\centering
\caption{Dice score comparison across the 12 source-to-target transfer pairs. Ours denotes the full dual co-training method, SHOT denotes the SHOT-style pseudo-label baseline, EMA denotes the EMA teacher baseline, FSM denotes the Fourier Style Mining baseline, UPL denotes the Uncertainty-aware Pseudo Label Guided baseline, and AIF denotes the Autonomous Information Filter-driven baseline.}
\label{tab:dice_results}
\begin{tabular}{cllcccccc}
\toprule
No. & Source & Target & Ours & SHOT & EMA & FSM & UPL & AIF \\
\midrule
1  & MTID   & TaL1   & 0.790 & 0.768 & \textbf{0.793} & 0.564 & 0.752 & 0.746 \\
2  & MTID   & UX2020 & \textbf{0.811} & 0.755 & 0.758 & 0.623 & 0.765 & 0.766 \\
3  & MTID   & UXTD   & \textbf{0.782} & 0.726 & 0.767 & 0.666 & 0.714 & 0.735 \\
4  & CTID   & Cleft  & \textbf{0.747} & 0.705 & 0.708 & 0.599 & 0.682 & 0.591 \\
5  & CTID   & UXSSD  & \textbf{0.801} & 0.771 & 0.776 & 0.712 & 0.748 & 0.731 \\
6  & UPX    & UXTD   & \textbf{0.802} & 0.771 & 0.778 & 0.702 & 0.723 & 0.727 \\
7  & UXTD   & UXSSD  & \textbf{0.822} & 0.784 & 0.799 & 0.759 & 0.773 & 0.760 \\
8  & Cleft  & UPX    & \textbf{0.737} & 0.659 & 0.672 & 0.628 & 0.608 & 0.624 \\
9  & TaL1   & UXTD   & 0.653 & 0.548 & 0.575 & 0.642 & \textbf{0.671} & 0.564 \\
10 & UX2020 & TaL1   & \textbf{0.71}6 & 0.587 & 0.650 & 0.535 & 0.576 & 0.579 \\
11 & UXTD   & MTID   & \textbf{0.710} & 0.644 & 0.691 & 0.703 & 0.686 & 0.547 \\
12 & UXSSD  & CTID   & \textbf{0.743} & 0.706 & 0.711 & 0.704 & 0.623 & 0.645 \\
\midrule
\multicolumn{3}{c}{\textbf{Average}} & \textbf{0.760} & 0.702 & 0.723 & 0.653 & 0.693 & 0.668 \\
\bottomrule
\end{tabular}
\end{table*}

\begin{table*}[t]
\centering
\caption{MSD (px) comparison across the 12 source-to-target transfer pairs. Ours denotes the full dual co-training method, SHOT denotes the SHOT-style pseudo-label baseline, EMA denotes the EMA teacher baseline, FSM denotes the Fourier Style Mining baseline, UPL denotes the Uncertainty-aware Pseudo Label Guided baseline, and AIF denotes the Autonomous Information Filter-driven baseline.}
\label{tab:msd_results}
\begin{tabular}{cllcccccc}
\toprule
No. & Source & Target & Ours & SHOT & EMA & FSM & UPL & AIF \\
\midrule
1  & MTID   & TaL1   & 1.902 & 2.136 & \textbf{1.810} & 9.117 & 2.145 & 2.018\\
2  & MTID   & UX2020 & \textbf{1.528} & 2.223 & 2.198 & 4.592 & 2.206 & 1.907\\
3  & MTID   & UXTD   & \textbf{2.453} & 3.188 & 2.486 & 3.413 & 5.372 & 3.683\\
4  & CTID   & Cleft  & \textbf{2.193} & 2.508 & 2.552 & 4.222 & 3.417 & 6.743\\
5  & CTID   & UXSSD  & \textbf{1.750} & 2.086 & 2.164 & 2.917 & 2.037 & 1.928 \\
6  & UPX    & UXTD   & \textbf{2.131} & 2.613 & 2.465 & 2.910 & 2.920 & 2.549\\
7  & UXTD   & UXSSD  & \textbf{1.386} & 1.852 & 1.602 & 1.898 & 1.699 & 1.759\\
8  & Cleft  & UPX    & \textbf{2.928} & 4.680 & 4.326 & 5.167 & 5.444 & 5.138\\
9  & TaL1   & UXTD   & 4.008 & 5.558 & 4.591 & \textbf{3.829} & 4.409 & 6.468\\
10 & UX2020 & TaL1   & \textbf{3.403} & 8.054 & 5.682 & 8.668 & 8.611 & 7.323\\
11 & UXTD   & MTID   & 3.132 & 4.660 & 3.649 & \textbf{2.834} & 5.017 & 6.772\\
12 & UXSSD  & CTID   & \textbf{2.128} & 2.888 & 2.655 & 2.573 & 4.461 & 3.481 \\
\midrule
\multicolumn{3}{c}{\textbf{Average}} & \textbf{2.412} & 3.537 & 3.015 & 4.345 & 3.978 & 4.147\\
\bottomrule
\end{tabular}
\end{table*}

\subsection{Ablation Study}
Table~\ref{tab:ablation_results} summarizes the effect of removing each component from the dual co-training framework. On average, the full model attains the lowest MSD (2.412 px) and highest Dice (0.760), while removing the EMA teacher, GAN refinement, or pseudo-label quality control consistently degrades both metrics to varying degrees. Across the 12 transfer pairs, the variants without EMA and without GAN refinement show moderate drops, but the QC ablation yields the largest reduction in Dice (down to 0.735 on average), indicating that filtering unreliable pseudo-labels is critical for maintaining overlap accuracy. The impact of EMA is particularly pronounced for the challenging UX2020$\rightarrow$TaL1 transfer, where removing it causes MSD to more than double and Dice to fall from 0.716 to 0.622, confirming its role in stabilizing adaptation under severe domain shift.

\begin{table*}[t]
\centering
\caption{Ablation study on the proposed dual co-training framework. Full denotes the full method, "EMA" removes the EMA teacher, "GAN FT" removes periodic GAN refinement, and "QC" removes pseudo-label quality control.}
\label{tab:ablation_results}
\begin{tabular}{cllcccc|cccc}
\toprule
 & & & \multicolumn{4}{c}{Dice$\uparrow$} & \multicolumn{4}{c}{MSD$\downarrow$} \\
\cmidrule(lr){4-7} \cmidrule(lr){8-11}
No. & Source & Target & Full & EMA & GAN FT & QC & Full & EMA & GAN FT & QC \\
\midrule
1  & MTID   & TaL1   & \textbf{0.790} & 0.786 & 0.789 & 0.780 & \textbf{1.902} & 1.918 & 1.949 & 2.134 \\
2  & MTID   & UX2020 & 0.811 & \textbf{0.812} & 0.807 & 0.733 & 1.528 & \textbf{1.524} & 1.536 & 2.619 \\
3  & MTID   & UXTD   & \textbf{0.782} & 0.777 & 0.779 & 0.763 & 2.453 & 2.650 & 2.517 & \textbf{2.266} \\
4  & CTID   & Cleft  & \textbf{0.747} & 0.740 & 0.734 & 0.721 & \textbf{2.193} & 2.268 & 2.407 & 2.281 \\
5  & CTID   & UXSSD  & \textbf{0.801} & 0.800 & 0.793 & 0.754 & \textbf{1.750} & 1.859 & 1.890 & 2.156 \\
6  & UPX    & UXTD   & \textbf{0.802} & 0.799 & 0.781 & 0.773 & \textbf{2.131} & 2.198 & 2.405 & 2.289 \\
7  & UXTD   & UXSSD  & \textbf{0.822} & 0.816 & 0.819 & 0.777 & \textbf{1.386} & 1.514 & 1.458 & 1.590 \\
8  & Cleft  & UPX    & \textbf{0.737} & 0.715 & 0.737 & 0.648 & \textbf{2.928} & 3.563 & 2.948 & 4.067 \\
9  & TaL1   & UXTD   & \textbf{0.653} & 0.615 & 0.596 & 0.644 & 4.008 & 4.304 & 4.700 & \textbf{3.815} \\
10 & UX2020 & TaL1   & \textbf{0.716} & 0.622 & 0.675 & \textbf{0.716} & \textbf{3.403} & 7.292 & 3.777 & 3.685 \\
11 & UXTD   & MTID   & 0.710 & 0.701 & 0.653 & \textbf{0.721} & 3.132 & 3.443 & 4.148 & \textbf{2.887} \\
12 & UXSSD  & CTID   & 0.743 & \textbf{0.744} & 0.743 & 0.732 & 2.128 & \textbf{2.124} & 2.578 & 2.182 \\
\midrule
\multicolumn{3}{c}{\textbf{Average}} & \textbf{0.760} & 0.744 & 0.743 & 0.735 & \textbf{2.412} & 2.888 & 2.693 & 2.664 \\
\bottomrule
\end{tabular}
\end{table*}

\begin{figure}
    \centering
    \includegraphics[width=1\linewidth]{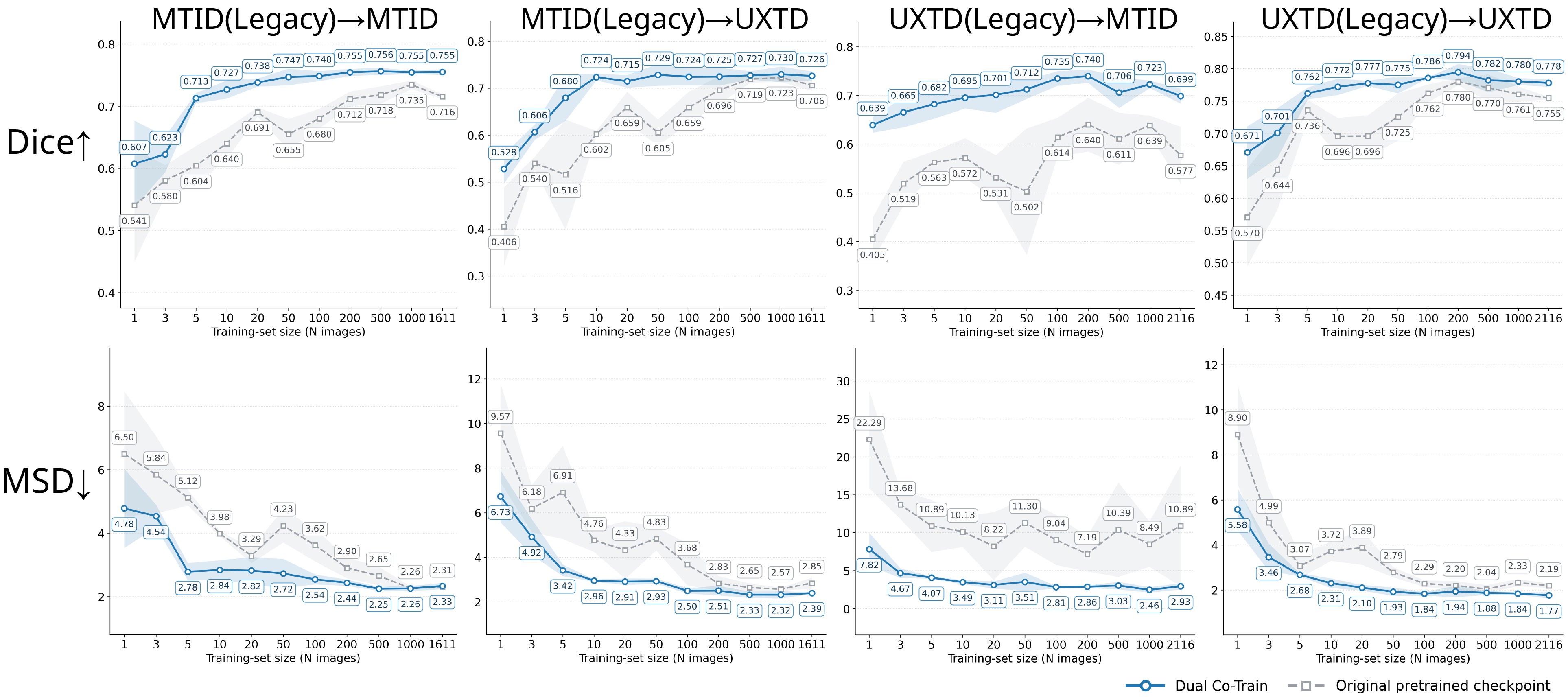}
    \caption{Source-size scaling comparison between the dual co-training framework and the plain pretrained checkpoint across four transfer configurations. Dice (top) and MSD (bottom) are reported as a function of training-set size $N$, averaged over three random seeds; shaded bands denote standard deviation. The dual co-training framework saturates at substantially smaller $N$ and exhibits lower variance across seeds than the baseline.}
    \label{fig:scaling}
\end{figure}

\subsection{Source-Size Scaling}

To assess how segmentation performance scales with the amount of labeled source data, we conduct a scaling study using two large legacy datasets from the original UltraUNet benchmark \cite{ultraunet}: UXTD (2,116 images) and MTID (1,611 images). For each dataset, we train the source segmenter on subsets of size $N \in \{1, 3, 5, 10, 20, 50, 100, 200, 500, 1000, N_{\max}\}$, where $N_{\max}$ is the full legacy dataset size, and evaluate both the plain pretrained checkpoint and the full dual co-training framework on four transfer configurations: MTID(Legacy)$\rightarrow$MTID, MTID(Legacy)$\rightarrow$UXTD, UXTD(Legacy)$\rightarrow$MTID, and UXTD(Legacy)$\rightarrow$UXTD. Each configuration is repeated three times with different random seeds for subset sampling, and we report the mean and standard deviation of Dice and MSD across seeds.

Fig.~\ref{fig:scaling} shows that both methods improve as $N$ increases, but the dual co-training framework saturates substantially earlier than the baseline. For MTID(Legacy)$\rightarrow$MTID, our method reaches 0.713 Dice at $N=5$, matching the baseline's performance at its full training set
of 1,611 images (0.716 Dice). Similarly, for UXTD(Legacy)$\rightarrow$MTID, our method already achieves 0.639 Dice at $N=1$, exceeding the baseline's peak of 0.640 Dice reached only at $N=200$. This pattern holds across all four configurations: the dual co-training framework consistently reaches within a few percentage points of its own asymptotic performance using 10-50 labeled source images, whereas the baseline requires several hundred to over a thousand images to approach comparable Dice and MSD values. The dual co-training framework also exhibits visibly narrower
variance bands across seeds at every training size, indicating lower sensitivity to which specific images are sampled for the source set, in contrast to the baseline's wider and less stable confidence intervals, particularly at small $N$.

Within-dataset transfers (MTID(Legacy)$\rightarrow$MTID and UXTD(Legacy)$\rightarrow$UXTD) reach higher final Dice scores (0.755 and 0.778, respectively) than cross-dataset transfers (MTID(Legacy)$\rightarrow$UXTD: 0.726; UXTD(Legacy)$\rightarrow$MTID: 0.699), consistent with the expectation that same-dataset transfer involves less domain shift than cross-dataset transfer.









\section{Discussion}

The results suggest that the proposed framework is particularly effective in low-label source-free ultrasound tongue segmentation, where domain shift is substantial, and both annotated target and any source data are unavailable. The performance improvement over the baseline is most pronounced in the low-data regime, as evidenced by the scaling experiments. 

A notable finding is that transfers from the relatively cleaner MTID and CTID datasets to more noise-corrupted targets such as TaL1 and Cleft often yielded better performance than expected. This trend is consistent with the idea that the domain-specific augmentation used in our framework helps the model adapt beyond simple intensity normalization and better capture target-style contour variability. In contrast, the transfer from noisy TaL1 to UXTD remained one of the most challenging scenarios, suggesting that severe appearance mismatch and annotation noise can still limit source-free adaptation even when pseudo-label refinement is used. Interestingly, the standard UltraUNet pipeline \cite{ultraunet}, which uses histogram matching to tackle the domain shift, has shown the opposite results: the segmentation model performed better when transferred from more noisy sources to cleaner ones. These observations support the view that ultrasound tongue contour segmentation benefits from an adaptation strategy that is aware of both imaging artifacts and contour structure, rather than relying only on generic domain adaptation recipes.

Another interesting result is that, in several transfers, the proposed model achieved MSD and Dice values comparable to or even better than the human inter-rater variability range. This does not imply that the model surpasses human annotation quality in a strict sense, but it does suggest that the segmenter may have learned a contour preference that is consistent with the dominant annotation style in the training source and transferable to the target domain. In practice, this may be useful for applications that require stable and reproducible contour tracking, especially when the target dataset is noisy or annotations are subjective. At the same time, the result should be interpreted cautiously, since inter-rater agreement itself varies across datasets and does not define a single absolute upper bound.

The scaling studies further show that the benefit of the proposed framework is largest when labeled source data are scarce, and that this benefit reflects a genuine gain in labeling efficiency rather than only improved transfer. In the MTID(Legacy)$\rightarrow$MTID configuration, our framework attains 0.713 Dice with only 5 labeled images, matching the plain supervised baseline's performance at its full training set of 1,611 images (0.716 Dice); similarly, for UXTD(Legacy)$\rightarrow$MTID, our method reaches 0.639 Dice with a single labeled image, exceeding the baseline's peak of 0.640 Dice reached only at $N=200$. This suggests that the closed-loop combination of pseudo-label refinement, quality control, and synthetic augmentation extracts substantially more signal per labeled image than standard supervised training, functioning as an annotation-efficient training strategy in its own right, independent of its cross-dataset transfer benefits.

The ablation study clarifies the role of each component. Removing periodic GAN refinement consistently degrades performance, indicating that refreshing the synthetic branch is important for keeping the generated image distribution aligned with the evolving target masks. Removing pseudo-label quality control causes the largest drop in Dice score, confirming that filtering unreliable masks is crucial for preventing noisy target predictions from propagating through self-training. Removing EMA teacher also reduced performance, but the impact was more moderate and varied across transfer pairs. This pattern suggests that the framework works best when pseudo-label cleaning, synthetic target-style augmentation, and teacher stabilization are used together rather than in isolation. The design is therefore best understood as a closed-loop adaptation system in which each branch reduces a different failure mode of source-free training.

A limitation of this study is that benchmarking against external SFDA methods is inherently difficult in the present setting. Many published methods are built around different backbones, loss functions, or training schedules, and several are tuned for larger pretrained models rather than the low-label UltraUNet regime \cite{medSegDAreview}. For this reason, faithful reimplementation would require substantial redesign and additional hyperparameter tuning, which makes direct comparison less controlled. For this reason, many tested baselines did not perform well under experimental conditions of extreme label scarcity and were therefore excluded.

Furthermore, the scaling results suggest that the dual co-training framework substantially reduces the number of labeled source images required to approach a dataset's achievable performance ceiling, often reaching comparable Dice within 10-50 images rather than several hundred. This supports the practical motivation of the framework: rather than requiring extensive source annotation, a small labeled set combined with pseudo-label refinement and synthetic augmentation can recover most of the achievable performance on a given target domain. Interestingly, the same-dataset transfer performance ceiling (around 0.755-0.778 Dice) is not substantially higher than several cross-dataset transfer ceilings, despite the absence of true domain shift. Since the legacy MTID and UXTD annotations were produced by a different set of annotators than the ones used for our held-out test sets, part of this saturation likely reflects inherent inter-rater annotation variability rather than a genuine model capacity limit. This suggests that, beyond a moderate amount of labeled data, further gains may be fundamentally bounded by annotation consistency rather than by model or algorithmic improvements, an important consideration when interpreting apparent performance ceilings in low-label segmentation benchmarks.

Overall, the main contribution of this work is not only the performance improvement itself, but also the demonstration that source-free ultrasound tongue segmentation can benefit from a task-specific adaptation loop that combines pseudo-label filtering, synthetic contour-conditioned augmentation, and teacher-student refinement. The results indicate that this design is especially useful when annotation is limited and domain shift is strong, both of which are common in real ultrasound tongue imaging studies. Future work could examine whether the same strategy transfers to additional tongue imaging corpora, more diverse annotation protocols, or lighter adaptation schemes that reduce the need for synthetic generation.

\section{Conclusion}
We presented a source-free dual co-training framework for cross-dataset ultrasound tongue contour segmentation under low-label conditions. By combining pseudo-label self-training, contour-based quality control, and segmentation-guided synthetic augmentation, the proposed method improves adaptation performance within the same lightweight UltraUNet framework while remaining practical for real-time use. Experiments across 12 transfer pairs, scaling studies, and ablation results show that the framework is effective in both cross-dataset and in-dataset settings, especially when labeled source data are limited.

The results also highlight the importance of task-specific adaptation for ultrasound tongue imaging, where domain shift, annotation variability, and limited supervision remain major barriers. Rather than relying on generic adaptation recipes, our approach leverages structural contour information and target-domain synthesis to better match the imaging characteristics of each transfer scenario. This makes the framework a promising option for reproducible and label-efficient tongue segmentation across different recording conditions.

Future work will explore extending the framework to additional ultrasound corpora, stronger uncertainty modeling for pseudo-label selection, and more efficient adaptation strategies that reduce the need for synthetic generation while preserving robustness under severe domain shift.

\section*{Acknowledgments}
The work was supported in part by the Research Grant Council of Hong Kong, under Grant GRF No.15217224. (Corresponding author: Prof. Yongping Zheng) and by the Research Institute for Smart Ageing (RISA). 

This work involved human subjects in its research and approval of all ethical and experimental procedures and protocols was granted by the Human Subjects Ethics Sub-Committee of Hong Kong Polytechnic University under application No. HSEARS20240327011 \& HSEARS20240306010.

The authors used Generative AI during the preparation of this manuscript. The tool was applied for minor language editing, grammar corrections, and improving overall clarity and readability. It was not used to generate new research hypotheses, interpret data, or formulate scientific conclusions. The authors maintain ultimate responsibility for the accuracy and originality of all content presented in this manuscript.

\bibliographystyle{unsrt}  
\bibliography{references}

\end{document}